\documentclass[letterpaper, 10 pt, conference]{ieeeconf}  

\IEEEoverridecommandlockouts                              

\usepackage{xcolor}  
\usepackage{graphics} 
\usepackage{graphicx}
\usepackage[utf8]{inputenc}
\usepackage{amssymb}
\usepackage{amsmath} 
\usepackage{multirow}
\usepackage{booktabs}
\usepackage{hyperref}
\usepackage{cite}
\usepackage{array}
\hypersetup{hypertex=true,
colorlinks=true,
linkcolor=blue,
anchorcolor=blue,
citecolor=blue}

\usepackage{geometry}
\title{\LARGE \bf
LSGS-Loc: Towards Robust 3DGS-Based Visual Localization for Large-Scale UAV Scenarios
}

\author{Xiang Zhang, Tengfei Wang, Fang Xu, Xin Wang and Zongqian Zhan
\thanks{This work was supported by the National Natural Science Foundation of China (No.42301507) }
\thanks{All the authors are with the School of Geodesy and Geomatics, Wuhan University, China PR. (Corresponding Author: Xin Wang, xwang@sgg.whu.edu.cn)
}%
}

\makeatletter

\newcommand{\Rmnum}[1]{\expandafter\@slowromancap\romannumeral #1@}
\makeatother

\begin{document}

\maketitle
\thispagestyle{empty}
\pagestyle{empty}

\begin{abstract}

Visual localization in large-scale UAV scenarios is a critical capability for autonomous systems, yet it remains challenging due to geometric complexity and environmental variations. While 3D Gaussian Splatting (3DGS) has emerged as a promising scene representation, existing 3DGS-based visual localization methods struggle with robust pose initialization and sensitivity to rendering artifacts in large-scale settings. To address these limitations, we propose \textbf{\textit{LSGS-Loc}}, a novel visual localization pipeline tailored for large-scale 3DGS scenes. Specifically, we introduce a scale-aware pose initialization strategy that combines scene-agnostic relative pose estimation with explicit 3DGS scale constraints, enabling geometrically grounded localization without scene-specific training. Furthermore, in the pose refinement, to mitigate the impact of reconstruction artifacts such as blur and floaters, we develop a Laplacian-based reliability masking mechanism that guides photometric refinement toward high-quality regions. Extensive experiments on large-scale UAV benchmarks demonstrate that our method achieves state-of-the-art accuracy and robustness for unordered image queries, significantly outperforming existing 3DGS-based approaches. Code is available at: \url{https://github.com/xzhang-z/LSGS-Loc}

\end{abstract}

\section{INTRODUCTION}


Visual localization aims to estimate the 6-DoF camera pose of a query image relative to a pre-constructed 3D scene representation. As a cornerstone of spatial perception, this technology underpins a wide array of applications, ranging from augmented and virtual reality (AR/VR) to autonomous driving and robotic navigation. In particular, with the rapid proliferation of unmanned aerial vehicles (UAVs) and autonomous aerial systems, achieving robust visual localization in large-scale environments has become a critical yet challenging research frontier.


Existing approaches for large-scale visual localization can be broadly categorized into two paradigms. The first relies on \textit{explicit geometric structures}, where the scene is modeled using sparse or dense 3D points associated with local image descriptors \cite{sarlin2019coarse}. In this framework, the camera pose is typically recovered by solving a Perspective-\textit{n}-Point (PnP) problem based on 2D-3D correspondences established via feature matching \cite{lowe2004distinctive}. While these structure-based methods can yield high precision under controlled conditions, their performance is intrinsically tied to the reliability of feature correspondences. In large-scale scenarios, factors such as weak textures, repetitive structures, significant viewpoint variations, and limited view overlap often lead to matching failures \cite{taira2018inloc}, thereby severely compromising localization robustness.


The second paradigm encompasses \textit{learning-based methods}, which implicitly encode scene priors into neural network weights. Representative techniques include Absolute Pose Regression (APR), which directly infers camera pose from a single image \cite{kendall2015posenet, chen2024map}, and Scene Coordinate Regression (SCR), which predicts dense 3D scene coordinates for image pixels to recover pose geometrically \cite{brachmann2023accelerated}. Despite notable progress, these data-driven approaches encounter distinct bottlenecks in large-scale settings. APR methods often lack the geometric rigor to achieve centimeter-level accuracy, whereas SCR methods typically require scene-specific retraining and tend to generalize poorly as scene scale and geometric complexity increase \cite{wang2024glace, jiang2025r}.

Recently, the advent of 3D Gaussian Splatting (3DGS)~\cite{kerbl20233d} has pioneered a new paradigm for visual localization. Unlike classical sparse geometric representations or purely implicit neural fields, 3DGS provides an explicit, differentiable, and high-fidelity scene representation, making it highly attractive for pose estimation. Existing 3DGS-based localization methods can be categorized into three primary strands: \textit{optimization-based methods}, which iteratively refine camera pose by minimizing photometric inconsistency between the query image and rendered views \cite{botashev2024gsloc}; \textit{feature-enhanced methods}, which attach high-dimensional descriptors to Gaussian primitives to establish dense correspondences for PnP-based localization \cite{huang2025sparse}; and \textit{hybrid geometric methods}, which lift 2D matches into 3D using rendered geometry and estimate pose through rigid transformation solvers \cite{ liu2024gs, lu20253dgs_lsr }. Although these approaches have demonstrated encouraging results, their applicability in large-scale environments is still hindered by two fundamental hurdles.


The first hurdle concerns \textbf{robust pose initialization}, which directly influences the convergence in the final pose refinement stage. In addition, a reliable initial pose must achieve both cross-scene generalization and absolute geometric grounding, yet existing methods often address only one of these requirements. Feature-based and SCR-based methods suffer from fragile correspondences or the need for scene-specific retraining \cite{brachmann2023accelerated, wang2024glace, sidorov2025gsplatloc}, while relative pose estimation networks, though offering better generalization, face scale ambiguity and unreliable translation estimates for global localization \cite{dong2025reloc3r}. Thus, obtaining a robust, geometrically grounded initial pose without scene-specific training remains a critical open problem.

The second hurdle lies in the \textbf{reliability of photometric pose refinement}. Most optimization-based 3DGS methods assume uniform trustworthiness across all rendered pixels \cite{fodor2025gs, botashev2024gsloc, zhou2024six, xin2024gauloc}. However, large-scale reconstructions often contain artifacts such as blur and floating artifacts due to sparse views and imperfect geometry. When these unreliable regions are treated equally during optimization, the resulting gradients may bias pose updates, leading to suboptimal convergence or local minima. Therefore, distinguishing reliable structural regions from low-quality content is essential for stable photometric refinement \cite{cheng2025logs}.

To address these limitations, we propose \textbf{LSGS-Loc}, a novel visual localization pipeline tailored for large-scale 3D Gaussian Splatting scenes. Our pipeline begins by retrieving database images that share co-visibility with the query image. Departing from conventional PnP-RANSAC procedures, we determine the initial pose by minimizing pixel-wise photometric error, effectively integrating a neural network-predicted relative pose with the explicit 3D scale constraints provided by 3DGS to resolve scale ambiguity. During pose refinement, to mitigate the impact of artifacts inherent in large-scale 3DGS reconstructions, we employ a Laplacian-driven reliability masking mechanism. This mechanism guides photometric-error optimization toward regions with higher rendering quality, thereby enabling accurate localization even under degraded rendering conditions. Our main contributions are as follows:
\begin{itemize}
    \item We propose a unified 3DGS-based visual localization framework for large-scale environments that explicitly addresses both robust pose initialization and reliable photometric refinement.
    \item A novel pose initialization strategy is introduced, which combines scene-agnostic relative pose estimation with explicit 3DGS scene-scale constraints, enabling geometrically grounded localization without scene-specific training.
    \item To address the visual localization degeneration due to blur, floaters, and rendering artifacts in large-scale 3DGS scenes, we develop a Laplacian-mask-guided photometric refinement method that significantly improves localization robustness.
\end{itemize}

\section{RELATED WORKS}
This section surveys the literature on visual localization. While traditional structure-based pipelines establish the foundational framework, recent advancements are primarily categorized into learning-based and neural representation-based methods.

\subsection{Learning-based Localization.} 
Through encoding scene representations within neural networks, learning-based methods achieve remarkable inference efficiency. These approaches are generally categorized into Absolute Pose Regression (APR) and Scene Coordinate Regression (SCR).

\noindent\textbf{Absolute Pose Regression (APR).} APR methods aim to establish a direct mapping from an input image to its 6-DoF camera pose \cite{kendall2015posenet, chen2024map}. Compared to structure-based methods, APR typically exhibits lower localization accuracy and struggles to generalize to viewpoints that deviate significantly from the training distribution. In large-scale scenes, the sparsity of training data further limits both the generalizability and precision of these models. Consequently, APR frameworks often function akin to sophisticated image retrieval systems. 

\noindent\textbf{Scene Coordinate Regression (SCR).} These methods predict the 3D scene coordinates for each pixel in a query image. By establishing dense 2D-3D correspondences, the camera pose is resolved using a robust PnP solver within a RANSAC loop \cite{brachmann2023accelerated, jiang2025r}. The constraints derived from these dense geometric correspondences enable SCR to surpass structure-based methods in terms of precision within small-scale indoor environments. However, in large-scale scenarios, the limited parameter capacity of neural networks hinders the representation of complex landscapes and intricate scene detail. Although some recent works leverage feature diffusion techniques to introduce the concept of co-visibility—thereby improving scalability in large environments, SCR still falls short of the high precision achieved by traditional structure-based pipelines \cite{wang2024glace}.

\subsection{Visual Localization with Neural Representations.} 

\noindent\textbf{NeRF-based Methods.} The advent of Neural Radiance Fields (NeRF)\cite{mildenhall2021nerf} in high-fidelity view synthesis has spurred the development of various localization frameworks. DFNet\cite{chen2022dfnet} leverages NeRF to synthesize diverse training images from novel viewpoints, thereby enhancing the generalization of APR. iNeRF\cite{yen2021inerf} pioneered pose estimation via rendering inversion, estimating camera poses by minimizing the photometric loss between rendered and observed pixels. To improve convergence stability, subsequent methods such as PNeRFLoc\cite{zhao2024pnerfloc} and NeRFMatch\cite{zhou2024nerfect} employ a coarse-to-fine strategy, refining pose based on coarse localization combined with specialized loss functions. Alternatively, methods including NeRFLoc\cite{liu2023nerf}, CROSSFIRE\cite{moreau2023crossfire}, and FQN\cite{germain2022feature} exploit NeRF to extract 3D descriptors or features at different positions to establish robust 2D-3D correspondences. Despite their potential, NeRF-based localization faces significant hurdles in large-scale environments with high-resolution imagery. First, the high computational cost associated with training and rendering renders pose optimization (e.g., iNeRF) prohibitively slow. Second, the limited model capacity of NeRF often leads to geometric blurring and visible artifacts when representing expansive scenes, which degrades the accuracy of photometric alignment and descriptor matching.

\noindent\textbf{3DGS-based Methods.} The emergence of 3D Gaussian Splatting (3DGS)\cite{kerbl20233d} has introduced a paradigm shift in visual localization, effectively overcoming the real-time rendering bottlenecks of NeRF. By employing explicit scene representation and tile-based rasterization, 3DGS achieves high-fidelity rendering at hundreds of frames per second (FPS). Current research leveraging 3DGS for localization can be categorized into three main streams: 

(1) \textit{Optimization-based Refinement}: Analogous to the iNeRF framework\cite{yen2021inerf}, these methods exploit the differentiable nature of 3DGS to optimize the 6-DoF camera pose by minimizing the photometric loss. Although 3DGS significantly accelerates this process, such methods remain highly sensitive to pose initialization. To mitigate  this, HGS-Loc\cite{niu2025hgsloc} introduces heuristic algorithms for optimal pose initialization, while GS-ReLoc\cite{fodor2025gs} and GSLoc\cite{botashev2024gsloc} incorporate a decaying Gaussian blur mechanism to expand the basin of convergence, albeit at the cost of increased computational overhead. Furthermore, LoGS\cite{cheng2025logs} enhances robustness by integrating image feature comparisons, yet they encounter significant bottlenecks in large-scale scenes. Six-DoF\cite{zhou2024six}and Gau-Loc\cite{xin2024gauloc} utilize the reprojection error from feature point matching to accelerate convergence; however, these methods remain constrained by the susceptibility of feature extraction to textureless environments. 

(2) \textit{Feature-enhanced Representations}: Recent works such as 6DGS\cite{matteo20246dgs} estimate poses by sampling rays on Gaussian surfaces, though their precision often lags behind alternative approaches. GSplatLoc\cite{sidorov2025gsplatloc}, STDLoc\cite{huang2025sparse} embed local feature descriptors directly into the Gaussian primitives to establish 2D-3D correspondences. However, these architectural modifications limit the generalizability of standard 3DGS models. GSVisLoc\cite{khatib2025gsvisloc} further trains additional generalized neural networks to predict the features of Gaussian primitives; nonetheless, in complex, large-scale UAV scenarios, selecting a sufficient number of valid primitives for 2D-3D matching from a massive pool remains a significant challenge. 

(3) \textit{Hybrid Geometric Methods}: Methods such as 3DGS-LSR\cite{lu20253dgs_lsr} render images at initial poses and utilize depth information to lift 2D matches into 3D space. However, the inherent challenges of feature matching under sparse views often manifest as poor rendering quality in 3DGS, thereby undermining the effectiveness of such pipelines. GS-CPR\cite{liu2024gs} employs MAST3R to establish dense 2D-3D correspondences, but this approach encounters difficulties when processing high-resolution imagery in expansive environments.

\begin{figure*}[t]
  \centering
  \includegraphics[trim={0 0 0 0},clip,width=\textwidth]{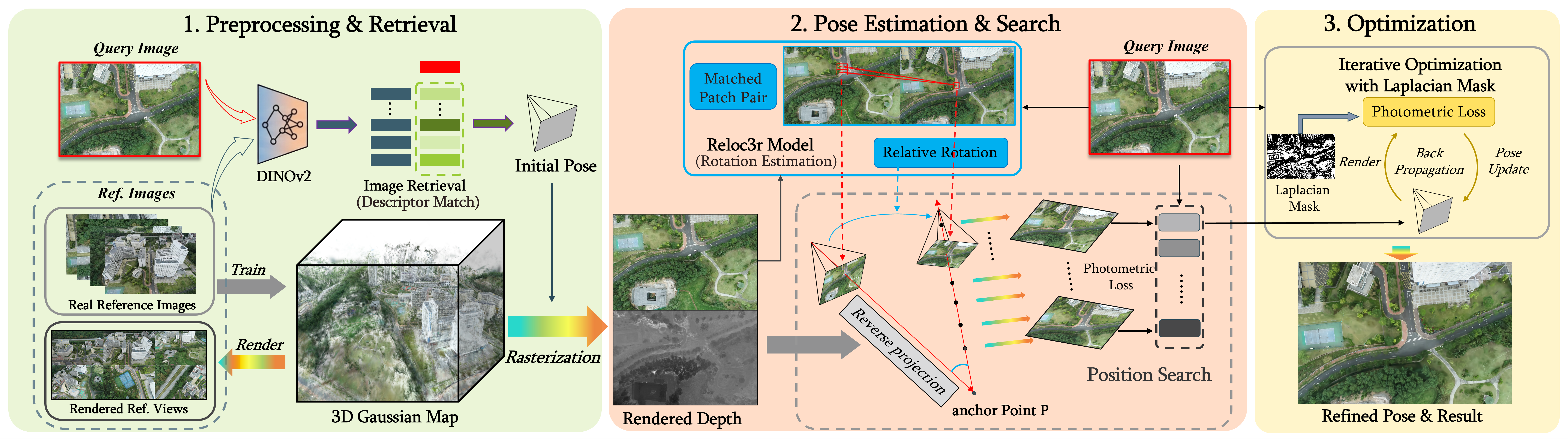}
  \caption{Workflow of the proposed LSGS-Loc. (1) Scene representation via 3DGS followed by reference retrieval; (2) Intermediate pose alignment based on relative pose estimation and depth-anchored photometric search; (3) Pose refinement using the Laplacian-mask-guided photometric loss.}
  \label{fig:2}
\end{figure*}

\section{METHODOLOGY}

\subsection{Overview of LSGS-Loc}

As illustrated in Fig. \ref{fig:2}, the proposed \textit{LSGS-Loc} framework consists of three sequential modules: preprocessing and retrieval, pose estimation and search, and iterative pose optimization.  The detailed workflow is described as follows:

\noindent\textbf{Phase 1: Preprocessing and Retrieval.} This stage establishes the scene representation by constructing a 3DGS model (more training details can be found at \cite{kerbl20233d}) from calibrated imagery. To enhance the robustness of retrieval and initial pose estimation, the reference database is augmented with extra synthetic views rendered from the 3DGS field. For a given query image, we employ the AnyLoc\cite{keetha2023anyloc} framework to identify the most similar reference frame within the augmented scene database. Based on the retrieved viewpoint, a corresponding synthetic RGB image and its associated depth map are rendered from the 3DGS representation to serve as the local reference for subsequent estimation.

\noindent\textbf{Phase 2: Pose Estimation and Search.} In this phase, the foundation model of ReLoc3r\cite{dong2025reloc3r} is leveraged to estimate the relative pose between the query image and the rendered reference. To resolve the inherent scale ambiguity associated with relative estimation, we extract internal attention maps to pinpoint the correspondence of the query's central patch within the rendered view. By leveraging the rendered depth, the 3D spatial position of this correspondence is determined via back-projection. This position serves as the origin for a photometric-guided search along the estimated orientation, establishing a reliable initial global pose.

\noindent\textbf{Phase 3: Pose Optimization.} The final phase refines the camera pose by minimizing the photometric loss between the query image and the 3DGS-rendered image. To mitigate the detriments of local rendering artifacts and geometric inconsistencies inherent to Gaussian primitives, we introduce a Laplacian-based masking mechanism. This mechanism selectively constrains the optimization process to image regions exhibiting high reconstruction fidelity and salient structural features, thereby ensuring superior robustness and convergence.

\subsection{Preprocessing and Retrieval}
Upon reconstructing the scene via 3D Gaussian Splatting (3DGS) using calibrated reference imagery, we construct a comprehensive database for visual retrieval. Following AnyLoc \cite{keetha2023anyloc} protocol, we extract compact global descriptors for all real reference images, enhancing robustness against complex urban structures and significant appearance variations.

To alleviate reduced observability caused by large occlusions (e.g., high-rise buildings and narrow streets), we augment the database with synthetic views rendered from novel viewpoints within the 3DGS field. Corresponding AnyLoc global descriptors are extracted for these rendered images and incorporated into the database alongside their associated camera poses. Consequently, the database comprises both real and synthetic reference views, ensuring denser and more comprehensive viewpoint coverage.

During inference, we extract a corresponding global descriptor for the query image and perform similarity retrieval across the database to obtain the most similar reference view and its camera pose. This retrieval outcome serves as both a local reference and a coarse pose initialization for subsequent fine-grained localization.

\subsection{Pose Estimation and Photometric-guided Search}
To obtain a robust initialization, based on the retrieved reference pose, we render an RGB image and its depth map from the 3DGS scene, forming a local reference observation. Then, the query image and rendered reference image are fed into the off-the-shelf ReLoc3r\cite{dong2025reloc3r} model to estimate the relative rotation $\Delta \mathbf{R}$ and relative translation direction $\Delta \mathbf{t}$. Acknowledging the scale ambiguity and potential instability of relative translation in challenging scenarios, we avoid simply converting $\Delta \mathbf{t}$ into a global position. Instead, we leverage internal attention-based coupled with rendered depth to explicitly guide the  camera position search.

Specifically, we extract cross-image attention responses from ReLoc3r to establish patch-to-patch correspondences. Given that high attention weights typically indicate corresponding physical regions, we identify the location of the query image's central patch within the rendered reference. By back-projecting this correspondence using the rendered depth map, we obtain a 3D anchor point $\mathbf{P}$, providing a metric-scale constraint.

First, we derive the query image's rotation $\hat{\mathbf{R}}_q$ by propagating the referenced rotation $\mathbf{R}_{ref}$ via the estimated relative rotation $\Delta \hat{\mathbf{R}}$ from ReLoc3r 
\begin{equation}
    \hat{\mathbf{R}}_q = \Delta \hat{\mathbf{R}} \cdot \mathbf{R}_{ref}
\end{equation}

Given the anchor $\mathbf{P}$ and its matched 2D points $\mathbf{(r,q)}$ on the retrieved and query images, $\Vec{q}$ is the vector from the camera center to $\mathbf{q}$ in the local camera framework. Then, query image's position search direction is supposed to be parallel to $\Vec{v} = \hat{\mathbf{R}}_q\Vec{q}$, corresponding to the world coordinate system. A 1D search is adopted starting from $\mathbf{P}$ along $-\Vec{v}$. The search interval $[d_{min}, d_{max}]$ is adaptively determined by scaling the rendered depth $d_P$ of the anchor point $\mathbf{P}$ as observed in the reference view:
\begin{equation}
    d_{min} = \gamma_{min} \cdot d_P, \quad d_{max} = \gamma_{max} \cdot d_P
\end{equation}
where $\gamma_{min}$ and $\gamma_{max}$ are configurable scaling coefficients. Given a fixed number of sample steps $Ns$, the step size is $\Delta d = (d_{max} - d_{min})/Ns$, forming a set of candidate positions:
\begin{equation}
    \mathcal{S} = \{ \mathbf{t}_k \mid \mathbf{t}_k = \mathbf{P} - (d_{min} + k \cdot \Delta d) \cdot \mathbf{v}, \, k \in [0, Ns] \}
\end{equation}

For each $\mathbf{t}_k$, we render a synthetic view and compute the $L_1$ photometric loss $\mathcal{L}(\mathbf{t}_k)$ against the query image. To mitigate convergence of local sub-optimal solutions, we leverage a discrete gradient $\nabla \mathcal{L}(\mathbf{t}_k)$ by averaging the absolute differences with adjacent views:

\begin{equation}
    \nabla \mathcal{L}(\mathbf{t}_k) = \frac{|\mathcal{L}(\mathbf{t}_k) - \mathcal{L}(\mathbf{t}_{k-1})| + |\mathcal{L}(\mathbf{t}_{k+1}) - \mathcal{L}(\mathbf{t}_k)|}{2}
\end{equation}

The optimal coarse translation $\mathbf{t}^{*}$ is then selected by jointly minimizing $\mathcal{L}(\mathbf{t}_k)$ and $ \nabla \mathcal{L}(\mathbf{t}_k)$:
\begin{equation}
\label{eq:min_search}
    \mathbf{t}^{*} = \arg \min_{\mathbf{t}_k} \left( \alpha \mathcal{L}(\mathbf{t}_k) + \beta \frac{1}{\nabla \mathcal{L}(\mathbf{t}_k) + \epsilon} \right)
\end{equation}
where $\alpha$ and $\beta$ are weighting parameters, and $\epsilon$ is a small constant for numerical stability. Finally, $\hat{\mathbf{R}}_q$ and $\mathbf{t}^{*}$ are used as an initialization for the subsequent pose refinement.

\subsection{Pose Optimization}
In the final stage, we further refine the full 6-DoF pose $\mathbf{T} \in SE(3)$ by minimizing the photometric discrepancy between the query image and the 3DGS-rendered view. At each iteration, we render an image from the 3DGS field under the current pose and compute the photometric loss to drive gradient-based refinement.

\noindent\textbf{Photometric objective.} 
Let $I_q$ denote the query image and $I_r(\mathbf{T})$ be the rendered image under pose $\mathbf{T}$. We minimize the $L_1$ photometric loss defined as:
\begin{equation}
\label{eq:photo_loss}
\mathcal{L}_{refine}(\mathbf{T}) = \frac{1}{|\Omega|} \sum_{\mathbf{u} \in \Omega} \| I_q(\mathbf{u}) - I_r(\mathbf{u}; \mathbf{T}) \|_1,
\end{equation}
where $\Omega$ represents the image domain.

\noindent\textbf{Lie algebra parameterization.} 
To ensure numerical stability and maintain the orthogonality of the rotation matrix during updates, we parameterize the incremental rotation using Lie algebra $\mathfrak{so}(3)$. Given an incremental vector $\boldsymbol{\phi} \in \mathbb{R}^3$, the rotation matrix $\mathbf{R}$ is updated via the exponential map:
\begin{equation}
\label{eq:rotation_update}
\mathbf{R} \leftarrow \exp([\boldsymbol{\phi}]_\times) \mathbf{R},
\end{equation}
where $[\boldsymbol{\phi}]_\times$ is the skew-symmetric matrix of $\boldsymbol{\phi}$.


\noindent\textbf{Laplacian-based mask.} 
To prevent initial poses near low-fidelity regions or 3DGS artifacts (e.g., floaters) from skewing optimization, we introduce an adaptive Laplacian-based mask. This mechanism exploits the correlation between high Laplacian responses and sharp textures, selectively suppressing contributions from blurry or structurally inconsistent areas during gradient descent.

\begin{table*}[t]
\centering
\caption{Comparison of different methods across tested datasets. Median rotation and translation errors ($^\circ$/cm). The results are conducted within two pools: (1) Original resolution methods, and (2) Resized/Pipeline-constrained methods. \textbf{Bold} and \underline{underline} indicate the best and second-best results within each pool.}
\vspace{-0.2cm}
\label{Comparison_Results}

\begin{tabular*}{1.0\linewidth}{@{\extracolsep{\fill}}llccccc}
\toprule
Method & Pipeline Res. & CUHK\_LOWER & CUHK\_UPPER & HAV & SMBU & SZIIT \\ \midrule
HLoc\cite{sarlin2019coarse} & original & \textbf{0.0137}/\underline{5.90} & \textbf{0.0137}/\underline{5.16} & \textbf{0.0084}/\underline{3.13} & \textbf{0.0091}/\textbf{3.66} & \textbf{0.0091}/\textbf{4.52} \\
\textbf{Ours} & original & \underline{0.0160}/\textbf{4.84} & \underline{0.0174}/\textbf{4.91} & \underline{0.0088}/\textbf{2.71} & \underline{0.0132}/\underline{3.87} & \underline{0.0166}/\underline{5.56} \\ \midrule

HLoc\cite{sarlin2019coarse} & 1600 & \textbf{0.0144}/6.13 & \textbf{0.0153}/6.40 & \textbf{0.0114}/3.76 & \textbf{0.0128}/5.23 & \textbf{0.0116}/\underline{5.68} \\
R-SCORE\cite{jiang2025r} & 722 & 0.2510/73.13 & 0.2208/77.74 & 0.1193/48.64 & 0.1092/50.46 & 0.3049/115.42 \\
ACE\cite{brachmann2023accelerated} & 722 & 0.21/62.87 & 0.23/81.51 & 0.17/53.51 & 0.16/75.18 & 0.22/83.32 \\
GLACE\cite{wang2024glace} & 722 & 0.28/84.97 & 0.23/84.21 & 0.17/57.41 & 0.17/75.59 & 0.25/97.36 \\
STDLoc\cite{huang2025sparse} & 1600 & 0.0936/26.87 & 0.0963/35.59 & 0.0780/23.23 & 0.0807/38.68 & 0.0779/32.75 \\
GS-CPR\cite{liu2024gs} & 1600 & 0.1841/60.10 & 0.2149/84.49 & 0.1436/51.51 & 0.1530/72.11 & 0.1909/87.14 \\
\textbf{Ours} & train:orig, test:1600 & \underline{0.0166}/\textbf{5.60} & 0.0181/\underline{5.82} & 0.0126/\textbf{3.17} & \underline{0.0147}/\textbf{4.63} & 0.0175/7.12 \\
\textbf{Ours} & 1600 & 0.0170/\underline{5.90} & \underline{0.0168}/\textbf{4.90} & \underline{0.0122}/\underline{3.43} & 0.0154/\underline{5.12} & \underline{0.0174}/\textbf{5.34} \\ \bottomrule
\end{tabular*}
\vspace{-4pt}
\end{table*}

During the first iteration, we partition the rendered image into $M$ non-overlapping patches $\{\mathcal{P}_j\}_{j=1}^{M}$ and compute a patch-wise Laplacian magnitude score $s_j$:
\begin{equation}
\label{eq:laplacian_score}
s_j = \frac{1}{|\mathcal{P}_j|} \sum_{\mathbf{u} \in \mathcal{P}_j} | \Delta I_r(\mathbf{u}; \mathbf{T}_0) |,
\end{equation}
where $\Delta$ is the discrete Laplacian operator and $\mathbf{T}_0$ is the initial pose. A binary mask is obtained $m(\mathbf{u})$ by thresholding:
\begin{equation}
\label{eq:binary_mask}
m(\mathbf{u}) = \mathbb{I} [ s_{\pi(\mathbf{u})} \ge \tau ],
\end{equation}
$\pi(\mathbf{u})$ maps pixel $\mathbf{u}$ to its corresponding patch index and $\tau$ is a predefined threshold. Then, only masked images are applied to Eq.~\eqref{eq:photo_loss} for all subsequent iterations to guide the optimization on high-confidence structural features.

\begin{table*}[t]
\centering
\caption{Ablation study of the multi-stage pose estimation pipeline. We report the pose recall (\%) under $(2^\circ, 20\text{m})$ and $(1^\circ, 10\text{m})$ thresholds for each phase configuration.}
\label{tab:ablation}
\resizebox{\textwidth}{!}{
\begin{tabular}{lcccccccccc}
\toprule
\multirow{2}{*}{Method} & \multicolumn{2}{c}{CUHK\_LOWER} & \multicolumn{2}{c}{CUHK\_UPPER} & \multicolumn{2}{c}{HAV} & \multicolumn{2}{c}{SMBU} & \multicolumn{2}{c}{SZIIT} \\
\cmidrule(lr){2-3} \cmidrule(lr){4-5} \cmidrule(lr){6-7} \cmidrule(lr){8-9} \cmidrule(lr){10-11}
 & 2$^\circ$/20m & 1$^\circ$/10m & 2$^\circ$/20m & 1$^\circ$/10m & 2$^\circ$/20m & 1$^\circ$/10m & 2$^\circ$/20m & 1$^\circ$/10m & 2$^\circ$/20m & 1$^\circ$/10m \\
\midrule
Phase 1     & 48.90 & 35.00 & 44.30 & 35.00 & 50.40 & 37.60 & 51.30 & 33.20 & 52.30 & 35.10 \\
Phase 1+2   & \textbf{97.30} & 90.10 & \textbf{98.70} & 87.80 & 97.90 & 89.40 & 97.90 & 87.70 & \textbf{98.30} & 88.90 \\
Phase 1+2+3 & 96.00 & \textbf{94.20} & \textbf{98.70} & \textbf{97.00} & \textbf{98.60} & \textbf{97.90} & \textbf{98.90} & \textbf{97.90} & 96.50 & \textbf{95.10} \\
\bottomrule
\end{tabular}
}
\end{table*}

\section{EXPERIMENT}

\subsection{Experimental Settings}

\noindent\textbf{Datasets.}
We evaluate on five UAV scenes from \textbf{GauUscene}~\cite{xiong2024gauu}: \texttt{CUHK\_LOWER}, \texttt{CUHK\_UPPER}, \texttt{HAV}, \texttt{SMBU}, and \texttt{SZIIT}. Following a 1:2 uniform temporal split, sampled frames constitute the \textbf{test set}, while the remainder reconstruct 3DGS maps using gsplat~\cite{gsplat}. 

\noindent\textbf{Implementation Details.}
During the retrieval phase, we augment the reference database via sequence densification, inserting one synthetic viewpoint at the geometric midpoint between consecutive real-world images. Global descriptors are extracted using the AnyLoc~\cite{keetha2023anyloc}, and the top-3 most similar images are retained as retrieval candidates. 

For the pose estimation and search phase, relative rotation $\Delta \mathbf{R}$ is predicted by a pre-trained ReLoc3r~\cite{dong2025reloc3r} backbone. Cross-attention maps from the sixth Transformer layer identify patch correspondences for geometric anchoring. A 1D photometric search samples adaptively within $[0.2 d_P, 2.0 d_P]$ ($N_s=30$), where $d_P$ is the rendered depth at anchor $\mathbf{P}$. We set $\alpha=0.8$, $\beta=0.2$ in Eq.~\eqref{eq:min_search}.

In the pose optimization phase, pose parameters (unit quaternions) are refined via Adam with an initial learning rate $0.018$ and cosine annealing, capped at 200 iterations. Our LSGS-Loc is implemented in PyTorch on a single NVIDIA RTX 4090 GPU.

\subsection{Comparisons with other SOTA methods}

We benchmark LSGS-Loc against state-of-the-art methods across three categories: 
(1) \textbf{Structure-based (SB)}. \textit{HLoc} \cite{sarlin2019coarse}, a hierarchical framework configured with SuperPoint~\cite{detone2018superpoint} and SuperGlue~\cite{sarlin2020superglue} for feature extraction and matching;
(2) \textbf{Scene Coordinate Regression (SCR) methods}. \textit{ACE}~\cite{brachmann2023accelerated}, \textit{GLACE}~\cite{wang2024glace}, and \textit{R-SCORE}~\cite{jiang2025r}; 
(3) \textbf{3DGS-based methods}. \textit{STDLoc}~\cite{huang2025sparse} (feature-enhanced) and \textit{GS-CPR}~\cite{liu2024gs} (hybrid geometric).

To ensure fairness, all learning-based baselines are retrained or fine-tuned on the same training split of the GauUscene dataset. Acknowledging distinct input constraints, we evaluate methods at their optimal supported scales. SCR methods (\textit{ACE, GLACE, R-SCORE}) are tested at their standard resolution (480px height, $\approx$722px width), as their network architectures are often optimized for specific receptive fields. 3DGS-based baselines (\textit{STDLoc, GS-CPR}) are evaluated at 1600px, the maximum resolution feasible within their original pipelines under the experimental setup. Our method is evaluated at both original and 1600px scales to facilitate comprehensive cross-resolution comparison.

\textbf{Quantitative Comparison on GauU-Scene Datasets.} 
Tab.~\ref{Comparison_Results} summarizes the median rotation ($^\circ$) and translation (\text{cm}) errors on the GauUscene dataset. Results are reported under two conditions: original resolution and downsampled resolution (1600px width), accommodating the constraints of various baseline pipelines.

Our proposed method achieves competitive performance across all five UAV scenes. Compared to the classical SB-based method \textit{HLoc}, our approach exhibits superior accuracy in translation estimation (e.g., achieving a median error of \textbf{2.71 cm} in the HAV scene vs. 3.13 cm for \textit{HLoc}). Although \textit{HLoc} maintains comparable rotation precision due to explicit geometric constraints, our method provides more balanced and robust localization in complex drone-view environments. Furthermore, our LSGS-Loc significantly outperforms coordinate regression-based and neural rendering-based baselines, such as ACE and GS-CPR, by a substantial margin. Specifically, while SCR-based methods (\textit{ACE, GLACE}) often suffer from accumulated drift in large-scale UAV trajectories (errors $>50$ cm), our method consistently maintains centimeter-level precision. This demonstrates that integrating 3DGS scene representation with iterative pose optimization effectively bridges the gap between efficiency and high-precision localization. Notably, comparing with 3DGS-based method, even at the downsampled 1600px resolution, our method retains higher accuracy, showcasing robustness to input variations and potential for practical deployment.

\begin{figure}[h]
\centering
\includegraphics[width=\linewidth]{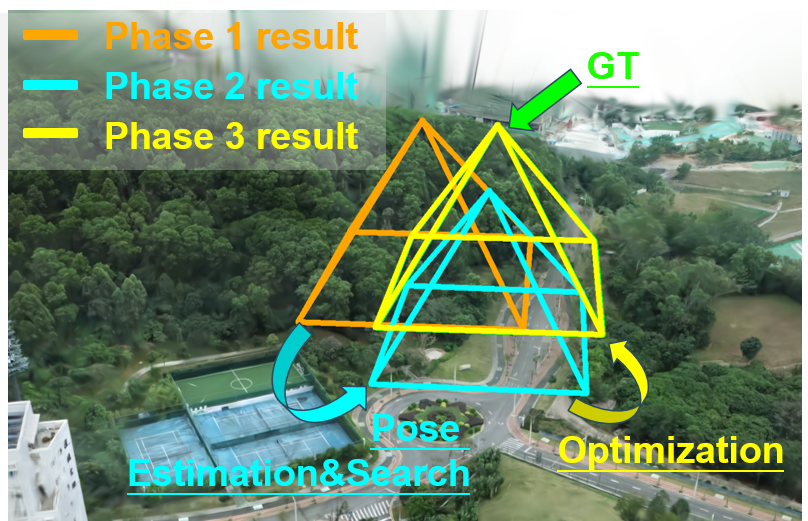} 
\caption{Visualization of the camera localization process. The Illustration of our progressive pose estimation aligned with Ground Truth (GT) through "retrieval", "Estimate \& Search", and "Optimization" stages.}
\label{fig:frustum_viz}
\end{figure}

\begin{figure}[t]
    \centering
    \includegraphics[width=\linewidth]{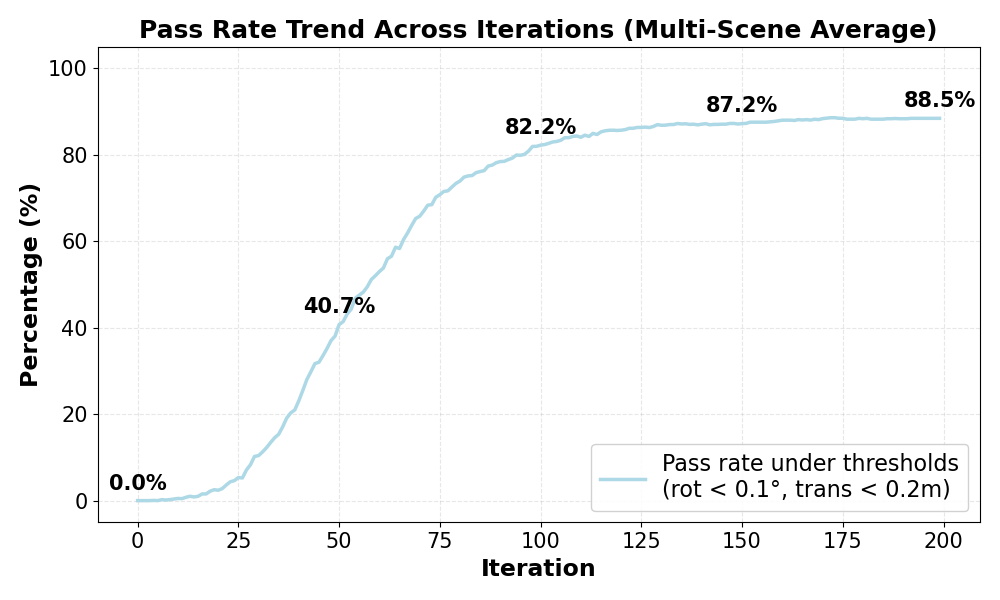}
    \caption{Pass rate under strict thresholds ($0.1^\circ$, $0.2$m) across 200 optimization iterations, averaged over all scenes.}
    \label{fig:iteration}
\end{figure}

\begin{figure*}[t] 
    \centering
    \includegraphics[width=\textwidth]{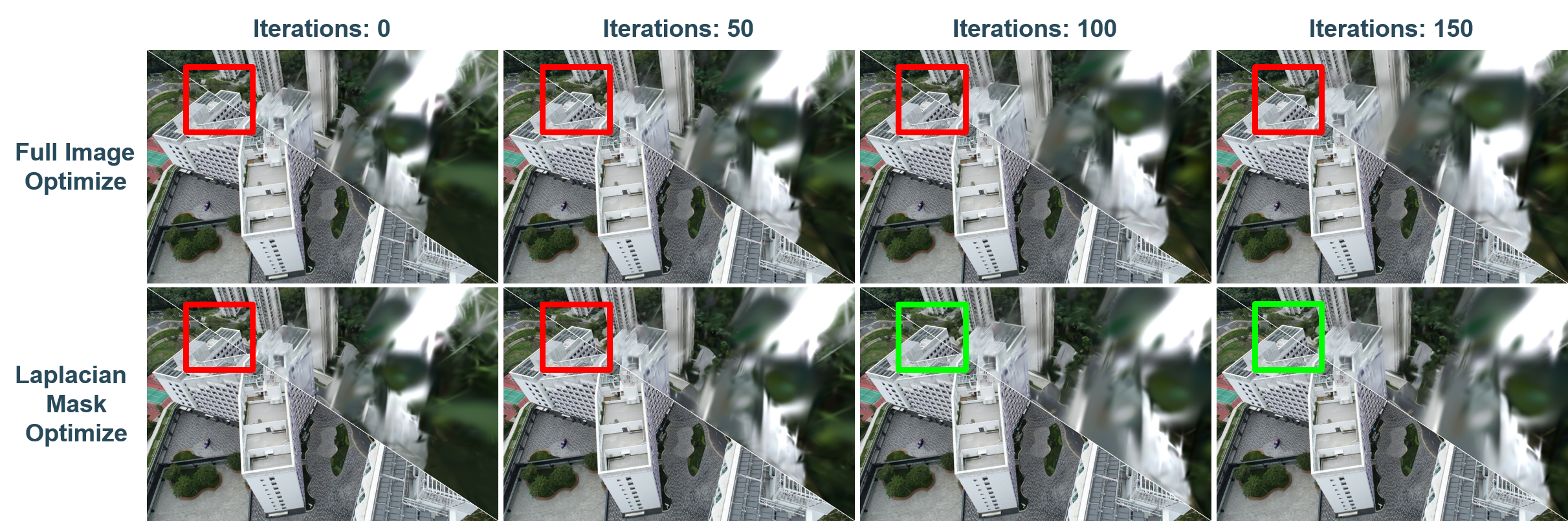} 
    \caption{Qualitative comparison of different optimization methods. The diagonal partitions the ground-truth query (lower-left) and the rendered image (upper-right).} 
    \label{fig:iter_compare} 
\end{figure*}

\subsection{Ablation Studies}
We conduct extensive ablation studies to validate the effectiveness of each component within the LSGS-Loc framework.

\textbf{Efficacy of Sequential Modules.}  For each stage, the results across five scenarios are shown in Tab.~\ref{tab:ablation}, summarizing the recall rates under ($2^\circ$, $20$m) and ($1^\circ$, $10$m) thresholds.

It can be found that Phase 1 only yields a coarse localization, with the ($1^\circ$, $10$m) recall limited to approximately 35\%. This limitation stems from the inherent viewpoint discrepancy between the query and the retrieved reference images. Integrating Phase 2 significantly improves precision: the ($2^\circ$, $20$m) recall exceeds 97\%, and the stricter ($1^\circ$, $10$m) recall increases by over 50 percentage points on average. This substantial gain confirms that the ReLoc3r-based estimation and our photometric-guided search effectively bridge the geometric gap and resolve the scale ambiguity, providing a high-quality initialization. In the final stage, Phase 3 further refines the pose, improving the ($1^\circ$, $10$\text{m}) recall to over 94\% while maintaining high performance at the looser threshold. This indicates that the optimization stage successfully converges most residual errors from Phase 2, underscoring the robustness of our complete pipeline. The qualitative results in Fig.~\ref{fig:frustum_viz} illustrate this progressive refinement. Notably, the ground truth frustum is occluded by the Phase 3 estimate due to their near-perfect alignment, highlighting the high-precision convergence of our method.

To analyze the convergence behavior of Phase 3, Fig.~\ref{fig:iteration} depicts the pass rate trend during the final pose optimization under a stringent threshold of ($0.1^\circ$, $0.2$m). The recall rate increases rapidly within the first 100 iterations, rising from 0\% to 82.2\%. The optimization eventually plateaus, achieving a final high-precision recall of 88.5\% at 200 iterations. This demonstrates the capability of Phase 3 to consistently refine the coarse initialization into high-accuracy poses.

\begin{table}[htbp]
\centering
\caption{Ablation results of the Laplacian-based mask.  The median rotation error ($^\circ$) and translation error (m) are reported.}
\label{tab:laplacian_ablation}
\begin{tabular}{llcc}
\toprule
Scene ($N$) & Method & Rot. ($^\circ$) $\downarrow$ & Trans. (m) $\downarrow$ \\
\midrule
\multirow{2}{*}{CUHK\_LOWER} & Full-image Opt. & 0.3706 & 0.9827 \\
 & \textbf{Laplacian Mask} & \textbf{0.0702} & \textbf{0.3743} \\
\midrule
\multirow{2}{*}{CUHK\_UPPER} & Full-image Opt. & 0.0655 & 0.3233 \\
 & \textbf{Laplacian Mask} & \textbf{0.0497} & \textbf{0.2655} \\
\midrule
\multirow{2}{*}{SZIIT} & Full-image Opt. & 0.0313 & 0.1960 \\
 & \textbf{Laplacian Mask} & \textbf{0.0296} & \textbf{0.1634} \\
\bottomrule
\end{tabular}
\end{table}

\begin{table*}[t]
\centering
\caption{Ablation study on the impact of the retrieval database augmentation. We report the median rotation error ($^\circ$) and median translation error (cm). ``DB Aug.'' denotes the database augmentation using rendered novel views.}
\label{tab:db_augmentation}
\resizebox{\textwidth}{!}{
\begin{tabular}{lccccc}
\toprule
Method & CUHK\_LOWER & CUHK\_UPPER & HAV & SMBU & SZIIT \\
\midrule
Ours (w/o DB Aug.) & 0.0160$^\circ$ / 4.84cm & 0.0174$^\circ$ / 4.91cm & \textbf{0.0088$^\circ$} / 2.71cm & 0.0132$^\circ$ / 3.87cm & 0.0166$^\circ$ / 5.56cm \\
Ours (w/ DB Aug.) & \textbf{0.0158$^\circ$} / \textbf{4.79cm} & \textbf{0.0173$^\circ$} / \textbf{4.48cm} & 0.0092$^\circ$ / \textbf{2.37cm} & \textbf{0.0122$^\circ$} / \textbf{3.86cm} & \textbf{0.0161$^\circ$} / \textbf{5.06cm} \\
\bottomrule
\end{tabular}
}
\end{table*}

\textbf{Efficacy of Laplacian-based Mask.} We evaluate the Laplacian-based mask against a full-image optimization baseline using a challenging subset of query images specifically selected from poorly reconstructed regions. As summarized in Tab.~\ref{tab:laplacian_ablation}, omitting the mask leads to a consistent increase in both rotation and translation median errors across all tested scenes, confirming its critical role in robust pose refinement. 

The effect is most pronounced in the CUHK\_LOWER scene, where the median translation error decreases from $0.9827$ m to $0.3743$ m, and the rotation error is reduced from $0.3706^\circ$ to $0.0702^\circ$ when the mask is applied. This improvement indicates that the Laplacian operator effectively identifies and suppresses regions with low reconstruction fidelity or 3DGS-related artifacts. By filtering these unreliable areas from the photometric loss computation, the optimization process is shielded from spurious gradients, leading to more stable and accurate convergence. 

Similar trends are observed in the \texttt{CUHK\_UPPER} and \texttt{SZIIT} scenes, where the mask consistently contributes to lower median errors. This cross-scene consistency validates the generality of our masking strategy in refining camera poses within explicit 3DGS representations. Beyond accuracy improvements, the Laplacian-based mask also enhances optimization stability. This is particularly evident in scenarios with large initial pose offsets or severe artifact occlusions, where the mask prevents the optimizer from converging to local minima or diverging entirely. As illustrated in Fig.~\ref{fig:iter_compare}, poses optimized with the Laplacian mask align closely with the ground truth, whereas the full-image optimization exhibits a noticeable drift due to unreliable photometric gradients from artifact-contaminated regions.


\textbf{Efficacy of Reference Database Augmentation.} Leveraging the novel view synthesis capability of 3DGS, we augment the retrieval database to enhance initialization quality. While standard evaluation protocols strictly utilize raw images to ensure fairness against baselines, this ablation study investigates the potential benefits of incorporating synthetic views. Specifically, we densify the trajectory by rendering novel views at geometric midpoints between temporally adjacent training images, effectively doubling the database size. 

To mitigate the feature domain gap between rendered and real images, which hinders direct comparison during retrieval, we process the top candidates from both image types independently during the Phase 2 pose search.

From Tab.~\ref{tab:db_augmentation}, this strategy consistently improves pose estimation accuracy across the tested scenes. This confirms that augmenting the retrieval database with rendered views provides denser viewpoint coverage and a more robust set of initializations for Phase 2, ultimately enhancing the overall localization precision.



\section{CONCLUSION}

In this paper, we propose LSGS-Loc, a novel large-scale visual localization framework based on 3DGS, tailored to address the key challenges of robust initialization and reliable refinement in large-scale environments. First, we introduce a robust pose initialization strategy that combines scene-agnostic relative pose estimation with explicit 3DGS scene-scale constraints, effectively utilizing the geometric and color information of 3DGS scenes to resolve scale ambiguity. Second, we propose a Laplacian-driven reliability masking mechanism for iterative photometric refinement. This mechanism selectively guides the optimization process toward regions with high reconstruction fidelity, ensuring stable convergence even under the interference of rendering artifacts and blur. The conducted experiments demonstrate that our LSGS-Loc successfully achieves centimeter-level positioning accuracy in complex unmanned aerial vehicle (UAV) environments. Compared to existing scene coordinate regression (SCR) and other 3DGS-based methods, LSGS-Loc exhibits superior flexibility concerning input image resolution and delivers more precise pose estimation in large-scale environments.


\bibliographystyle{IEEEtran.bst}
\bibliography{IEEEabrv, references}

\begin{thebibliography}{10}
\providecommand{\url}[1]{#1}
\csname url@rmstyle\endcsname
\providecommand{\newblock}{\relax}
\providecommand{\bibinfo}[2]{#2}
\providecommand\BIBentrySTDinterwordspacing{\spaceskip=0pt\relax}
\providecommand\BIBentryALTinterwordstretchfactor{4}
\providecommand\BIBentryALTinterwordspacing{\spaceskip=\fontdimen2\font plus
\BIBentryALTinterwordstretchfactor\fontdimen3\font minus \fontdimen4\font\relax}
\providecommand\BIBforeignlanguage[2]{{%
\expandafter\ifx\csname l@#1\endcsname\relax
\typeout{** WARNING: IEEEtran.bst: No hyphenation pattern has been}%
\typeout{** loaded for the language `#1'. Using the pattern for}%
\typeout{** the default language instead.}%
\else
\language=\csname l@#1\endcsname
\fi
#2}}

\bibitem{1}
Y.~Chen, Z.~Chen, C.~Zhang, \emph{et~al.}, ``Gaussianeditor: Swift and controllable 3d editing with gaussian splatting,'' in \emph{Proceedings of the IEEE/CVF conference on computer vision and pattern recognition}, 2024, pp. 21\,476--21\,485.

\bibitem{2}
Z.~Ma and S.~Liu, ``A review of 3d reconstruction techniques in civil engineering and their applications,'' \emph{Advanced Engineering Informatics}, vol.~37, pp. 163--174, 2018.

\bibitem{3}
B.~Kerbl, G.~Kopanas, T.~Leimk{\"u}hler, and G.~Drettakis, ``3d gaussian splatting for real-time radiance field rendering.'' \emph{ACM Trans. Graph.}, vol.~42, no.~4, pp. 139--1, 2023.

\bibitem{4}
P.~Papantonakis, G.~Kopanas, B.~Kerbl, A.~Lanvin, and G.~Drettakis, ``Reducing the memory footprint of 3d gaussian splatting,'' \emph{Proceedings of the ACM on Computer Graphics and Interactive Techniques}, vol.~7, no.~1, pp. 1--17, 2024.

\bibitem{5}
X.~Wang, R.~Yi, and L.~Ma, ``Adr-gaussian: Accelerating gaussian splatting with adaptive radius,'' in \emph{SIGGRAPH Asia 2024 Conference Papers}, 2024, pp. 1--10.

\bibitem{6}
W.~Liu, T.~Guan, B.~Zhu, \emph{et~al.}, ``Efficientgs: Streamlining gaussian splatting for large-scale high-resolution scene representation,'' \emph{IEEE MultiMedia}, 2025.

\bibitem{7}
Z.~Fan, K.~Wang, K.~Wen, Z.~Zhu, D.~Xu, Z.~Wang, \emph{et~al.}, ``Lightgaussian: Unbounded 3d gaussian compression with 15x reduction and 200+ fps,'' \emph{Advances in neural information processing systems}, vol.~37, pp. 140\,138--140\,158, 2025.

\bibitem{8}
J.~C. Lee, D.~Rho, X.~Sun, J.~H. Ko, and E.~Park, ``Compact 3d gaussian representation for radiance field,'' in \emph{Proceedings of the IEEE/CVF Conference on Computer Vision and Pattern Recognition}, 2024, pp. 21\,719--21\,728.

\bibitem{9}
S.~S. Mallick, R.~Goel, B.~Kerbl, \emph{et~al.}, ``Taming 3dgs: High-quality radiance fields with limited resources,'' in \emph{SIGGRAPH Asia 2024 Conference Papers}, 2024, pp. 1--11.

\bibitem{10}
B.~Li, S.~Chen, L.~Wang, \emph{et~al.}, ``Retinags: Scalable training for dense scene rendering with billion-scale 3d gaussians,'' \emph{arXiv preprint arXiv:2406.11836}, 2024.

\bibitem{11}
Y.~Chen and G.~H. Lee, ``Dogs: Distributed-oriented gaussian splatting for large-scale 3d reconstruction via gaussian consensus,'' \emph{Advances in Neural Information Processing Systems}, vol.~37, pp. 34\,487--34\,512, 2025.

\bibitem{12}
S.~Durvasula, A.~Zhao, F.~Chen, \emph{et~al.}, ``Distwar: Fast differentiable rendering on raster-based rendering pipelines,'' \emph{arXiv preprint arXiv:2401.05345}, 2023.

\bibitem{13}
J.~Jo, H.~Kim, and J.~Park, ``Identifying unnecessary 3d gaussians using clustering for fast rendering of 3d gaussian splatting,'' \emph{arXiv preprint arXiv:2402.13827}, 2024.

\bibitem{14}
J.~Lee, S.~Lee, J.~Lee, J.~Park, and J.~Sim, ``Gscore: Efficient radiance field rendering via architectural support for 3d gaussian splatting,'' in \emph{Proceedings of the 29th ACM International Conference on Architectural Support for Programming Languages and Operating Systems, Volume 3}, 2024, pp. 497--511.

\bibitem{15}
K.~Cheng, X.~Long, K.~Yang, \emph{et~al.}, ``Gaussianpro: 3d gaussian splatting with progressive propagation,'' in \emph{Forty-first International Conference on Machine Learning}, 2024.

\bibitem{16}
J.~Zhang, F.~Zhan, M.~Xu, S.~Lu, and E.~Xing, ``Fregs: 3d gaussian splatting with progressive frequency regularization,'' in \emph{Proceedings of the IEEE/CVF Conference on Computer Vision and Pattern Recognition}, 2024, pp. 21\,424--21\,433.

\bibitem{17}
Z.~Yu, A.~Chen, B.~Huang, T.~Sattler, and A.~Geiger, ``Mip-splatting: Alias-free 3d gaussian splatting,'' in \emph{Proceedings of the IEEE/CVF conference on computer vision and pattern recognition}, 2024, pp. 19\,447--19\,456.

\bibitem{18}
Y.~Jiang, J.~Tu, Y.~Liu, \emph{et~al.}, ``Gaussianshader: 3d gaussian splatting with shading functions for reflective surfaces,'' in \emph{Proceedings of the IEEE/CVF Conference on Computer Vision and Pattern Recognition}, 2024, pp. 5322--5332.

\bibitem{19}
X.~Wu, J.~Xu, C.~Wang, \emph{et~al.}, ``Local gaussian density mixtures for unstructured lumigraph rendering,'' in \emph{SIGGRAPH Asia 2024 Conference Papers}, 2024, pp. 1--11.

\bibitem{20}
Z.~Peng, T.~Shao, Y.~Liu, \emph{et~al.}, ``Rtg-slam: Real-time 3d reconstruction at scale using gaussian splatting,'' in \emph{ACM SIGGRAPH 2024 Conference Papers}, 2024, pp. 1--11.

\bibitem{21}
H.~Matsuki, R.~Murai, P.~H. Kelly, and A.~J. Davison, ``Gaussian splatting slam,'' in \emph{Proceedings of the IEEE/CVF Conference on Computer Vision and Pattern Recognition}, 2024, pp. 18\,039--18\,048.

\bibitem{22}
Y.~Bao, T.~Ding, J.~Huo, \emph{et~al.}, ``3d gaussian splatting: Survey, technologies, challenges, and opportunities,'' \emph{IEEE Transactions on Circuits and Systems for Video Technology}, 2025.

\bibitem{23}
M.~Levoy and P.~Hanrahan, ``Light field rendering,'' in \emph{Seminal Graphics Papers: Pushing the Boundaries, Volume 2}, 2023, pp. 441--452.

\bibitem{24}
A.~Davis, M.~Levoy, and F.~Durand, ``Unstructured light fields,'' in \emph{Computer Graphics Forum}, vol.~31, no. 2pt1.\hskip 1em plus 0.5em minus 0.4em\relax Wiley Online Library, 2012, pp. 305--314.

\bibitem{25}
S.~J. Gortler, R.~Grzeszczuk, R.~Szeliski, and M.~F. Cohen, ``The lumigraph,'' in \emph{Seminal Graphics Papers: Pushing the Boundaries, Volume 2}, 2023, pp. 453--464.

\bibitem{26}
C.~Buehler, M.~Bosse, L.~McMillan, S.~Gortler, and M.~Cohen, ``Unstructured lumigraph rendering,'' in \emph{Proceedings of the 28th annual conference on Computer graphics and interactive techniques}, 2001, pp. 425--432.

\bibitem{27}
S.~E. Chen and L.~Williams, ``View interpolation for image synthesis,'' in \emph{Seminal Graphics Papers: Pushing the Boundaries, Volume 2}, 2023, pp. 423--432.

\bibitem{28}
B.~Mildenhall, P.~P. Srinivasan, M.~Tancik, J.~T. Barron, R.~Ramamoorthi, and R.~Ng, ``Nerf: Representing scenes as neural radiance fields for view synthesis,'' \emph{Communications of the ACM}, vol.~65, no.~1, pp. 99--106, 2021.

\bibitem{29}
T.~M{\"u}ller, A.~Evans, C.~Schied, and A.~Keller, ``Instant neural graphics primitives with a multiresolution hash encoding,'' \emph{ACM transactions on graphics (TOG)}, vol.~41, no.~4, pp. 1--15, 2022.

\bibitem{30}
Z.~Shu, R.~Yi, Y.~Meng, Y.~Wu, and L.~Ma, ``Rt-octree: accelerate plenoctree rendering with batched regular tracking and neural denoising for real-time neural radiance fields,'' in \emph{SIGGRAPH Asia 2023 Conference Papers}, 2023, pp. 1--11.

\bibitem{31}
J.~T. Barron, B.~Mildenhall, D.~Verbin, P.~P. Srinivasan, and P.~Hedman, ``Mip-nerf 360: Unbounded anti-aliased neural radiance fields,'' in \emph{Proceedings of the IEEE/CVF conference on computer vision and pattern recognition}, 2022, pp. 5470--5479.

\bibitem{32}
H.~Turki, D.~Ramanan, and M.~Satyanarayanan, ``Mega-nerf: Scalable construction of large-scale nerfs for virtual fly-throughs,'' in \emph{Proceedings of the IEEE/CVF conference on computer vision and pattern recognition}, 2022, pp. 12\,922--12\,931.

\bibitem{33}
X.~Cao, B.~Lin, B.~Wang, \emph{et~al.}, ``Ssnerf: Sparse view semi-supervised neural radiance fields with augmentation,'' \emph{arXiv preprint arXiv:2408.09144}, 2024.

\bibitem{34}
H.~Xiong, \emph{SparseGS: Real-time 360° sparse view synthesis using Gaussian splatting}.\hskip 1em plus 0.5em minus 0.4em\relax University of California, Los Angeles, 2024.

\bibitem{39}
J.~L. Schonberger and J.-M. Frahm, ``Structure-from-motion revisited,'' in \emph{Proceedings of the IEEE conference on computer vision and pattern recognition}, 2016, pp. 4104--4113.

\bibitem{41}
Z.~Zhan, R.~Xia, Y.~Yu, \emph{et~al.}, ``On-the-fly sfm: What you capture is what you get,'' \emph{ISPRS Annals of the Photogrammetry, Remote Sensing and Spatial Information Sciences}, vol. X-1-2024, pp. 297--304, 2024.

\bibitem{42}
J.~T. Barron, B.~Mildenhall, D.~Verbin, P.~P. Srinivasan, and P.~Hedman, ``Mip-nerf 360: Unbounded anti-aliased neural radiance fields,'' in \emph{Proceedings of the IEEE/CVF conference on computer vision and pattern recognition}, 2022, pp. 5470--5479.

\bibitem{43}
Z.~Zhan, Y.~Yu, R.~Xia, \emph{et~al.}, ``Sfm on-the-fly: Get better 3d from what you capture,'' \emph{arXiv preprint arXiv:2407.03939}, 2024.

\bibitem{44}
Z.~Zhu, S.~Peng, V.~Larsson, \emph{et~al.}, ``Nice-slam: Neural implicit scalable encoding for slam,'' in \emph{Proceedings of the IEEE/CVF conference on computer vision and pattern recognition}, 2022, pp. 12\,786--12\,796.

\bibitem{45}
X.~Yang, H.~Li, H.~Zhai, \emph{et~al.}, ``Vox-fusion: Dense tracking and mapping with voxel-based neural implicit representation,'' in \emph{2022 IEEE International Symposium on Mixed and Augmented Reality (ISMAR)}.\hskip 1em plus 0.5em minus 0.4em\relax IEEE, 2022, pp. 499--507.

\bibitem{46}
R.-W. Li, W.~Ke, D.~Li, L.~Tian, and E.~Barsoum, ``Monogs++: Fast and accurate monocular rgb gaussian slam,'' in \emph{35th British Machine Vision Conference 2024, {BMVC} 2024, Glasgow, UK, November 25-28, 2024}.\hskip 1em plus 0.5em minus 0.4em\relax BMVA, 2024.

\bibitem{47}
J.~Sturm, N.~Engelhard, F.~Endres, W.~Burgard, and D.~Cremers, ``A benchmark for the evaluation of rgb-d slam systems,'' in \emph{2012 IEEE/RSJ international conference on intelligent robots and systems}.\hskip 1em plus 0.5em minus 0.4em\relax IEEE, 2012, pp. 573--580.

\bibitem{48}
Y.~A. Malkov and D.~A. Yashunin, ``Efficient and robust approximate nearest neighbor search using hierarchical navigable small world graphs,'' \emph{IEEE transactions on pattern analysis and machine intelligence}, vol.~42, no.~4, pp. 824--836, 2018.

\end{thebibliography}

\end{document}